\RequirePackage{fix-cm}
\documentclass[smallextended]{svjour3}       
\smartqed  
\usepackage{graphicx}
\usepackage{comment}
\usepackage{cite}
\usepackage{marvosym}
\usepackage{amsmath}
\usepackage{algorithm}
\usepackage{algpseudocode}
\usepackage{marvosym}
\usepackage{afterpage}
\begin{document}

\title{Filling the Holes on 3D Heritage Object Surface\\based on Automatic Segmentation Algorithm}
\author{Sinh Van Nguyen \Letter \and Son Thanh Le \and Minh Khai Tran \and Le Thanh Sach}
\institute{Sinh Van Nguyen \Letter, Son Thanh Le, Minh Khai Tran \at
              School of Computer Science and Engineering \\
              International University - VNUHCM, Vietnam\\
              \email{nvsinh@hcmiu.edu.vn; ltson,tkminh@hcmiu.edu.vn}   \\
           Sach Thanh Le \at
           School of Computer Science and Engineering \\
           Ho Chi Minh City University of Technology - VNUHCM, Vietnam\\
              \email{ltsach@hcmut.edu.vn} 
}
\date{Received: date / Accepted: date}
\maketitle
\begin{abstract}
Reconstructing and processing the 3D objects are popular activities in the research field of computer graphics, image processing and computer vision. The 3D objects are processed based on the methods like geometric modeling (a branch of applied mathematics and computational geometry) or the machine learning algorithms based on image processing. The computation of geometrical objects includes processing the curves and surfaces, subdivision, simplification, meshing, holes filling, reconstructing and refining the 3D surface's objects on both point cloud data and triangular mesh. While the machine learning methods are developed using deep learning models. With the support of 3D laser scan devices and LiDAR techniques, the obtained dataset is close to original's shape of the real objects. Besides, the photography and its application based on the modern techniques in recent years help us collect data and process the 3D models more precise. This article propose an improved method for filling holes on the 3D object's surface based on an automatic segmentation. Instead of filling the hole directly as the existing methods, we now subdivide the hole before filling it. The hole is first determined and segmented automatically based on computation of its local curvature. It is then filled on each part of the hole to match its local curvature shape. The method can work on both 3D point cloud surfaces and triangular mesh surface. Comparing to the state-of-the-art (SOTA) methods, our proposed method obtained higher accuracy of the reconstructed 3D objects.

\keywords{Point clouds \and 3D Mesh \and Reconstruction \and Segmentation \and Hole-filling \and Digital heritage \and Virtual museum}

\end{abstract}

\section{Introduction}
The development of information technology (IT) in recent years makes our works changed in almost of fields. IT is used as a tool and also an important solution to support processing our work faster and more accurate. Reconstructing the 3D heritages objects is always interested not only by the archaeologist but also the scientist in computer science. In practice, the external reasons such as time, history, weather, natural disasters and even some reasons caused by human like war, etc., that made the cultural heritages are not preserved completely. How to preserve and restore them such that they are as close to their original shapes as possible. They are important duties and regarded by the researchers in different scientific fields. Geometric modeling is mathematical-based methods for processing, building and reconstructing 3D object models. It is used to compute the geometrical objects and their characteristics such as point, line, polyline, triangle, normal vector, curve and surface, etc., on both 2D and 3D environment \cite{Jonh2019}. They are background and widely used in mathematical functions applied in the field of computer graphics \cite{Steve2021}. For example, starting from the two points, a line is created; three points, a polygon or a triangle is formed; while for given four points, we can create a rectangle in 2D or a tetrahedron in 3D.\\
Reconstructing the 3D surfaces from unorganized point clouds or a triangular mesh data has interested by the researchers in simulated computation. The game developer spent most of time in creating 3D objects and processing their interactions in the games. The mathematicians focused on their studies to reconstruct, reshape and refine the 3D objects based on the mathematical equations. The architects and constructors are always working on 3D design sketch or drawings. While the archaeologists would like to restore and reshape the heritage samples as their original models \cite{Minh2021}. Although the high-tech devices like 3D scanner, 3D printer or LIDAR data acquisition techniques are emerged as major and modern tools to support researchers. They are used not only for obtaining data, but also for calibrating the samples to measure and obtain exactly their accuracy \cite{revos2023}. However, the final results depends on the proposed methods and techniques to collect, scan and acquire data for reconstructing the models.\\
The problem comes from samples status, the shapes complexity of real objects and also the different data acquisition techniques. In some cases, the density of data points are distributed irregularly on the object's surface through collecting process. This lead to creating some holes on the surface. In other cases, the samples may have lost some parts of their original shapes (e.g. a part of head, a piece of hands or legs, etc., of the history statues or a part of them was missing). Besides, the materials of samples have effected directly to the obtained datasets. The factors like smoothness, color of the object's surface; or structure and complexity of the object's surface, etc. Additional features to support increasing the quality of obtained datasets are data acquisition techniques of technicians. If the dataset collected with high quality, the next data processing step is faster and the reconstructed objects will have higher accuracy.\\
In this article, we propose a method for filling holes on the object's surface automatically. Our method is an improved step of the previous work \cite{SinhJ2022}. For each hole, we first compute and segment it based on its local curvature shape (i.e. at the locations of highest or lowest points/vertices on the hole boundary) to split the hole. The filling process is then filled on each segmented part of the hole. Comparing to the SOTA methods, our proposed method obtained higher accuracy. After filing the holes, the reconstructed objects are approximately fixed to their original shapes. At the end, the reconstructed objects of cultural heritages (e.g. the statues in the Ho Chi Minh Museum \cite{hcmmuseum}) are exhibited in the virtual museum based on VR and AR applications.\\   
The remaining of paper is structured as follows. We present introduction of our research project and scientific basis (e.g. mathematical concepts, definitions and notations, the theory of curve and surface, the subdivision and segmentation) in section 2. Section 3 reviews the related works in the SOTA methods. Our proposed method is presented in detail in section 4. Section 5 will discuss, evaluate and compare experiment results between our proposed method and previous work. The last section, section 6 is our conclusion.

\section{Introduction to research project and scientific basis}
This article is a part of our research project in digitalization and visualization of museum. A virtual museum is constructed from the historical museum of Vietnam. The description in detail of our project and necessary knowledge bases, even so the mathematical foundation, utility tools and techniques are presented as follows. 
\subsection{The research project DS2023-28-01}
In this section, we introduce about our research project identified by the name ``DS2023-28-01". This research project is approved by the scientific committee of Vietnam National University Ho Chi Minh City (VNU-HCM), in the field of computer science, with title ``A research for reconstruction, digitalization and visualization of tangible heritages in 3D space of VR and AR". It is funded by VNU-HCM, under grant number DS2023-28-01 and started at early of 2023. In order to study and implement the project, we have signed a research cooperation agreement with the Museum of Ho Chi Minh City (HCMC) \cite{hcmmuseum}, and the Museum of HCMC assigned an expert, as a partner joining with our research group to provide data source and historical information of these specimens. The Museum of HCMC is also well-known as a visiting place in HCMC Vietnam, where store and preserve from hundreds to thousand of historical objects. These tangible cultural heritage artifacts, specimens related to history, culture, people, antiques, etc., and are known with a cultural and historical thickness of over 3000 years; from the stone age, bronze age, pottery from the brilliant Oc Eo culture to the achievements in the process of reclamation and establishment of the hamlet of Vietnam. Vietnamese residents in the land of the South.\\
At first, we select a group of 17 worship statues to process (see Fig.1). For each of them, we collect data by scanning, capturing surround the statue and producing a training dataset for the next reconstructing and visualization steps.
\begin{figure}
\center
\includegraphics[width=\textwidth]{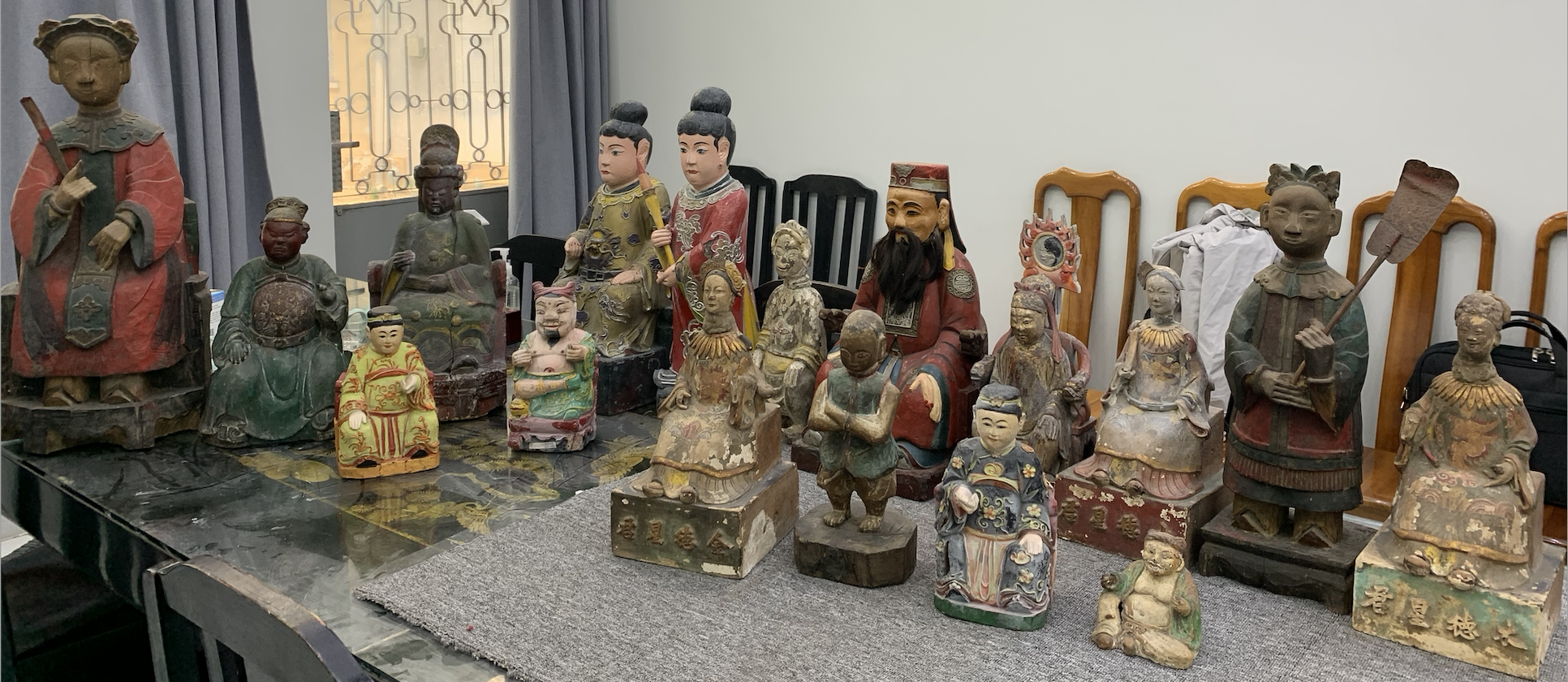}
\caption{The samples of heritages at Museum of HCMC \cite{hcmmuseum}.}
\end{figure}  
\noindent At present, this group of statues belongs to the same a topic and they are exhibited in the same room of the museum. After processing, they will also be exhibited in the same room of the virtual museum on a VR application. Our research project will be repeated in each exhibited room until the whole museum of HCMC is digitalized and visualized completely.
\subsection{Concepts and definitions}
As mentioned above, geometric modeling (or computational geometry) is used to process the geometrical objects in creating, reconstructing a 2D or a 3D object. The process on 2D objects handle the 2D components like a point, like, polyline, rectangle, circle, ellipse or a triangle. While the 3D objects processing includes a sphere, cylinder, cube, tetrahedron or any type of a 3D shape. We use the following concepts and mathematical notations in this research (in context of the 3D space) and our computation.
\begin{itemize}
\item[-] A ring of hole boundary on the 3D point cloud surface includes the 3D point such that lack of neighboring points at a projected plane of its 8-connectivity; while a ring of triangular hole boundary on the 3D triangular surface include the triangles such that each of these triangles existed an edge that is not shared with other (see Fig.2)
\item[-] A 3D surface shape is called an open surface if it contain a surface boundary (e.g. a surface of terrain or an elevation surface). In the contrary, a 3D surface shape is called a closed surface when it is closed from at all directions, but it contains a volume and may contains the hole boundary on its surface.
\item[-] $p_i$ a 3D point, with $P_i$ is a set of $p_i$
\item[-] $p_c$ a 3D central point
\item[-] $p_b$ a 3D boundary point
\item[-] $p_{nei}$ a neighbor 3D point
\item[-] $p_{new}$ a new 3D inserted point
\item[-] $p_{seg}$ a new 3D inserted point on the fracture margin
\item[-] $\overrightarrow{p_b}$ vector of $p_b$
\item[-] $e_b$ a boundary edge
\item[-] $R(e_b)$ a ring of boundary edges ($R_f(e_b)$: first ring, $R_n(e_b)$: next ring)
\item[-] $t_b$ a boundary triangle
\end{itemize}
The mathematical notations and concepts that we used in our researches are also presented in the previous work \cite{NVS2013}. 
\begin{figure}
\center
\includegraphics[width=\textwidth]{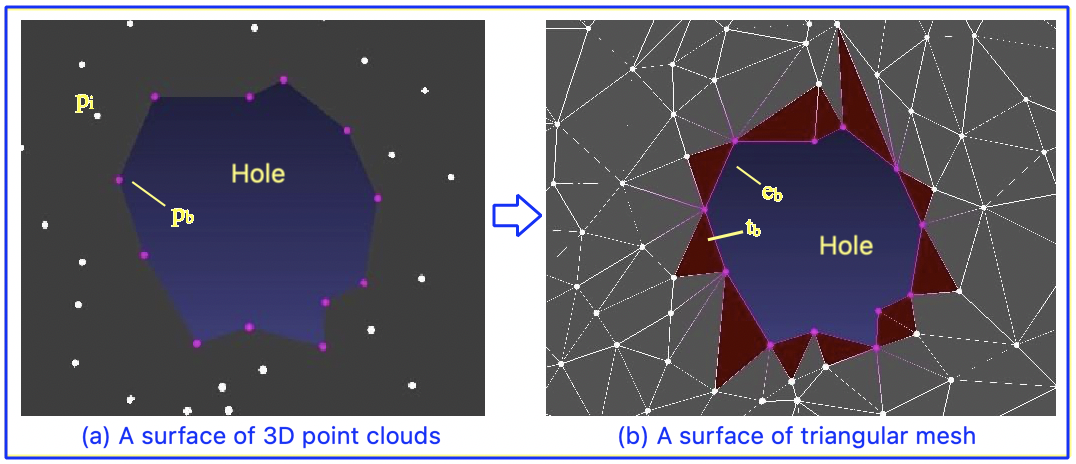}
\caption{The boundary of a hole (red color): (a) a hole boundary ring on the surface of 3D point clouds; (b) a hole boundary ring on the surface of 3D triangular mesh.}
\end{figure}  
\subsection{Curves and surfaces}
Theory of curve and surface is background for building geometrical objects and creating their 3D models \cite{Marco2012, Penev2013}. In order to create a curve, we need a set of points and connect them to obtain a polyline. The curve is then generated by interpolating their control points on the polyline. From the curve in 2D, the surface of a 3D object is also computed and created from a set of 3D points. To recall the formula of Bezier curve and surface as mentioned in \cite{David2001} and computed the reverse engineering of Bezier curve in \cite{Sinh2018}, the Bezier curve is discovered by the French engineer Pierre Bézier \cite{Licio2013}. It approximate tangents by using control points to generate curve and represent mathematically as follows:
\begin{equation}
	C(t)=\sum_{i=0}^{n}P_{i}B_{i}^{n}(t)
\end{equation}
\noindent where $0 \leq t \leq 1$, $P_i$ is values of control points between the two endpoints, $B_{i}^{n}(t)$ is computed as: $B_{i}^{n}(t) = \binom{n}{i}t^{i}(1-t)^{n-i}$, while $\binom{n}{i}$ is computed as $\frac{n!}{i!(n-i)!}$; Therefore, $C(t)$ at the end is presented as follows:
\begin{equation}	
	C(t)=\sum_{i=0}^{n}\frac{n!}{i!(n-i)!}t^{i}(1-t)^{n-i}P_{i}
\end{equation}
\noindent By the same way of computation, the Bezier surface is considered as a set of Bezier curves following the curvature of the surface. Its formula is presented as follows:
\begin{equation}
	Q(u,w)=\sum_{i=0}^{n}\sum_{j=0}^{m}B_{i,j}J_{i}^{n}(u)K_{j}^{m}(w)
\end{equation}
where $B_{i,j}$ are vertices of the defining polygon net, while $J_{i}^{n}(u)$ and $K_{j}^{m}(w)$ are computed as $B_{i}^{n}(t)$.
The Bezier surface has properties that we can apply to compute and process the 3D surface objects in practice as follows: 
\begin{itemize}
	\item[-] The degree of the surface in each polynomial direction is one less than the number of defining polygon vertices in that direction (e.g. degree 3 has 4 points, degree 4 has 5 points, etc).
	\item[-] The continuity of the surface in each parametric direction is two less than the number of defining polygon vertices
	\item[-] The surface generally follows the shape of the defining polygon net (i.e. the curvature of the surface is based on the positions of the vertices in each area of the surface).
	\item[-] The surface is contained in the convex hull and also in the concave of the polygon net depending on the shape of surface.
	\item[-] The surface is invariant under an affine transformation. 
	\item[-] Each of the boundary curves in this case is a Bezier curve that may belong to the set of Bezier curves of the surface.
\end{itemize}
The application of Bezier curve and surface is used to process the surface of a 3D object. In previous research work \cite{SHM2018,Sinh2018}, we applied them to fill the holes on the surface of 3D point clouds and the surface of a triangular mesh. By computing boundary points and their extended boundary points of the hole which generated new points to interpolate and insert into the hole region. The method has proved better the several methods with small holes on the surface. However, in case of larger holes with complex shape, it is difficult to fill the holes and reconstruct the surface. That lead to reason we suggest a method to segment a large and complex hole first; and then filling in each part of the hole to preserve the local curvature of hole surface and approximates to its original shape. 
\subsection{Segmentation}
Segmentation is a well-known method in image processing. Instead of processing in the whole area of picture, it is segmented into multiple parts. The analysis and handle for the next step is then processed on each part of the image. In geometric modeling \cite{Ergun2021}, the structures of real 3D objects obtained by scanning their surface. The quality of surface shape depends on scanning step and surface structure of the real objects. Because of many reasons caused such as point location, point density, surface triangulation of point clouds and even coming from the data acquisition process. Therefore, geometric segmentation is understood as subdividing (or spliting) the object's surface depend on its geometrical characteristics. As mentioned in \cite{Luca2015}, segmentation is an important task including semantic segmentation and geometric segmentation. The semantic segment try to subdivide the object into meaningful pieces to serve for labeling or processing based on machine learning. While the geometric segmentation aims at finding suitable location of points on the object surface to divide mesh or point clouds into clusters for next individual processing step. This idea is also proposed in the research of Gen Li et.al. \cite{Gen2008}. The method splits a complex and large hole into simple ones. After that, each simple part of the hole is filled with planar triangulation. This method is work well with the complex hole on the object surface. However, in case if the subdivided part is not flat, the algorithm may not filled adapting local curvature of the hole. In our method, we segment a large hole based on the second criterion (geometric segmentation), before filling and reconstructing it.
\section{Related works}
In this section, we study and review the SOTA methods to reconstruct the shapes of 3D objects or 3D models from different datasets. The several methods and their application in practice based on geometric modeling and machine learning applied in computer graphics, images processing and computer vision. These methods implemented based on two directions: hole filling on the 3D points clouds and hole filling on the triangular mesh. For each of them, there are also two approaches to process: volume-based methods and surface-based methods. We focus on the surface-based methods to reconstruct the shape of 3D real objects. 
\subsection{The geometric-based methods}
Luca Calatroni et.al. \cite{Luca2023} presented a geometric proposed framework for restoring a triangular surface of the 3D objects. The data obtained by scanning the real objects, and then the object's surface is triangulated for the next processing step. The framework introduced a list of tasks in geometric processing like removing noise scanned data \cite{Sinh2013}, filling holes \cite{Chunhong2017}, reconstructing the 3D meshes \cite{Sinh2014,Sinh2019}, etc., An application is then implemented and performed three main tasks (denoising, inpainting/hole filling and completion). By processing directly on the mesh surface, the application shown all computation of geometrical features on the object's surface. However, the large and complex holes in different curvature is not mentioned in the paper. To the small holes, they can easy to fill by using the SOTA methods.
In computer graphics, we compute the shapes of 3D objects based on geometric modeling. The holes on the object's surface is determine based on the hole boundary. If the data object is a point cloud, we process by using the method \cite{Sinh2016}; to the mesh surface, we handle based on the method \cite{Sinh2018}. After filling the holes, the object's surface is reconstructed adapting their local curvature on the surface. The extended topics in geometric modeling, image processing and computer graphics are described in detail in \cite{SinhJ2021}. This article summarized the proposed methods and application that are widely used and applied in practice. However, in reconstructing the surface of 3D objects, we are facing the problem of loosing, fault and/or lack of information that is constrained on the objects surface, and the density of points is distributed not regularly caused by the data collecting step.\\
Jules et. al. \cite{Jules2018} presented a new method for reconstructing the surface of unorganized point clouds. The method is based on a Poisson schema approximated the basis functions CSRBF (Compactly Supported Radial Basis Function). The method provided a good estimation of missing data in the scanning point cloud dataset to serve for restoring the forest objects. The method is processed using geometrical algorithms and has proved effectively of applied mathematics in computation.   
Xiaoyuan Guo at. al. \cite{Guo2016} presented a survey research of the geometric-based methods for filling the hole on the 3D objects of unorganized points set and mesh. The paper described in detail the SOTA methods in both theory analysis and experimental results to compare. This research provided us a full picture of the hole filling methods based on the geometric modeling computation.
Emiliano Perez et. al. \cite{Perez2016} presented a comparison of hole filling methods in 3D models. This is a deep research reviewing 34 SOTA methods and compare them together to figure out the advantages and also limitations for each method. The contribution of this paper provided valuable experiences to the researchers who are working in this field. A. L. Custodio et. al. \cite{Custodio2023} presented a mathematics-based method for filling the hole on the surface that approach with $C^1-smoothness$. The method is completely used mathematics under the two specific non-linear constraints: having a prescribed area and prescribed values of the mean curvature at some points. The quadratic splines is related to the fact that increasing the degree or the smoothness of the fitting splines; while the area constraint is considered the Bezier techniques that will approximate the quadratic Powell-Sabin splines by means of triangular patches. The obtained results proved the advantages comparing to the SOTA methods. Linyan Cui et.al. \cite{Cui2021} presented a geometrical-based method for detecting and repairing the hole on the surface of a 3D point cloud of symmetrical object. The method includes many steps: hole detection, hole boundary determination, hole filling and reconstruction. However, these steps presented in the paper are introduced and processed in the existing method \cite{SinhJ2021}; and this method is only worked well on the mechanical claw holes.  
The idea of decompose a complex hole into multiple simple holes to process is suggested in the research \cite{Gou2022}. Author segment a large hole based on hole boundary detection and edge extension computation. Each simple hole is then filled to obtain the hole fitting after repairing step. However, the results of proposed method is not analyzed and compared the quality measurement to the SOTA method. Therefore, it is difficult to evaluate exactly the method based on the visualization of picture.      

\subsection{The learning-based methods}
The methods for reconstructing or restoring the real 3D objects based on machine learning techniques are trends in recent researches. The main condition to process the 3D real objects using deep learning models is training dataset. With the rapid development if power server with GPU support, the time processing is not important nowadays. However, the input training dataset is most requires in any deep learning models. Bin Liu et. al. \cite{Bin2021} studied a method for surface segmentation based on learning local geometric features (e.g. the local curvature on the surface). The method is implemented including many steps: processing the 3D obtained model of utensil fragments based on their neighborhood geometric characteristics. To build a training dataset of features mapping images after labeling the ground-truth data. Working on the surface of triangular mesh to eliminate the mislabeled and residual triangle cells to obtain the fracture surface. This method proved effectively at extracting the core features of geometric data of utensil fragment.\\
Wenfeng Du et.al. \cite{Du2023} proposed a method for generating 3D solid models by applying a 3D GAN and reverse engineering (RE) technology based on 3D grid volume of voxels. After creating an initial model of the object, it is processed based on a deep learning model to produce an optimized object in voxels. The data model is then reconstructed both surface and 3D solid model based on RE. The obtained model is a completed solid 3D object. The combination of machine learning methods and geometric methods in this study can improve accuracy of the 3D model. This research can also be the input for industrial design and manufacture, especially for 3D printing technology. The technique of using GAN to process 3D from 2D image dataset is also proposed and processed efficiently in this research \cite{Weizhi2019}. The researched group of Alok Jhaldiyal et. al. \cite{Alok2023} presented a deep review in the SOTA methods covering topics in 3D point clouds processing that includes obtaining and registration of raw point data, classification, object detection, tracking, segmentation, transformation, extraction and reconstruction of 3D object based on the deep learning methods. The research topic is then focused on projected-based methods of 3D point clouds on the 2D plane for processing and their benefit for deep learning techniques. The analysis, comparison and discussion in the experimental results of the existing methods (to show their advantages and disadvantages) are main contributions of the paper in this field. Chen Chen et. al. \cite{Chen2022} suggested a method for semantic segmenting the 3D point clouds dataset. The method is based on encoder and processing boundary-guided decoder in which not only effectively exploits the geometric information of the whole object, but also explores the boundary area. The geometric features are analyzed such as boundary contrast learning and geometric representation like computation of density, distance between the points, color representation, etc., to indicate relation of machine learning-based method and geometric-based methods. The experimental results shown effectiveness of the proposed method comparing to the existing methods. 

\section{Proposed method}
Extending from the previous research work \cite{SinhJ2022}, we try to process and fill in a larger and complex hole on the object's surface. Normally, the hole is generated during scanning step. The reason may come from the scanning techniques or missing information of the real object as mentioned above. Depending on shape and structure of the real objects, a hole can be created at the hidden areas, on the local surface, ridge, concave or fracture slots, etc., of the surface. To the small hole, we can easily to process as previous work. To a larger hole or the hole with more complex curvature, the method must be performed such that it is fixed along with its shape after filling. In any way, after determining each region of the hole, region growing approach from the boundary points is an idea to create new points for inserting and filling the hole.
\begin{figure}[H]
\center
\includegraphics[width=\textwidth]{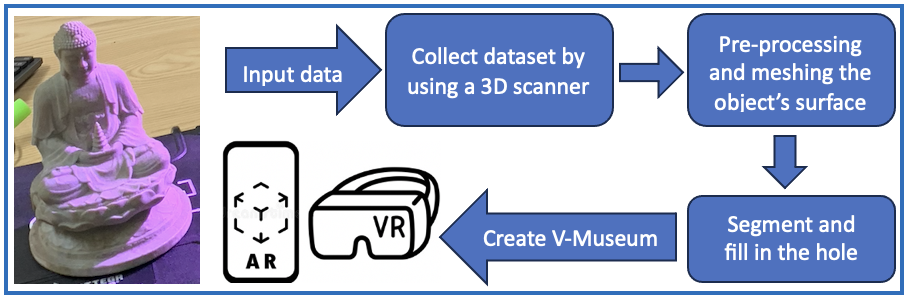}
\caption{Our proposed method for reconstructing the 3D object of heritage: (i) Collecting dataset; (ii) Pre-processing and meshing the object's surface; (iii) Segmenting and filling the holes on the surface; (iv) Building a V-museum application.}
\end{figure}  
\noindent The real data is obtained by using a 3D scanner. After that, it is pre-processed by removing noisy data and meshing the object's surface. The most important step is focused on hole filling step. After reconstructing the shape to approach initial model of the real objects, they are used and visualized in a V-museum application (see Fig.3).   
\subsection{Data obtaining and processing}
In this section, we describe our method to collect data. Depending on the shapes of the real objects, we use different 3D scanner devices. The samples in the museum of HCMC includes statues within the height of 1 meter (see Fig.1), we use two 3D scanners (see Fig.4). The scanner EinScan Pro+ is used to scan simple statues very fast. It support color processing. To the complex surface on the statues, we use the R-EVO scanner. This device support laser scanning and sensor to calibrate and measure the accuracy of the object's model.
\begin{figure}[H]
\center
\includegraphics[width=\textwidth]{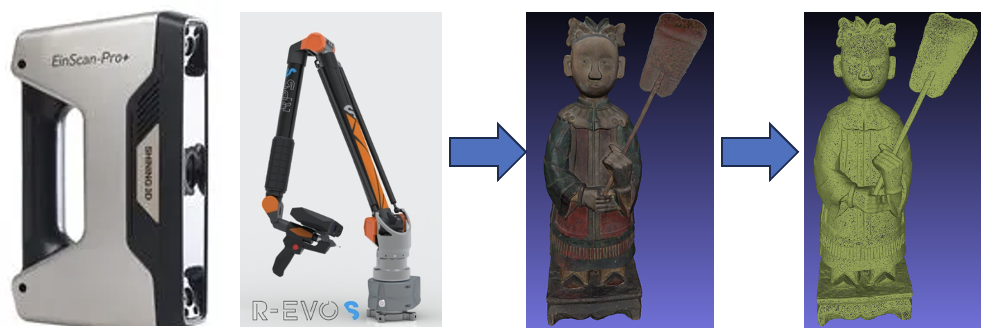}
\caption{Our method to collect and process data. From left to right: we use 3D scanners to scan data of the real statues in the museum of HCMC. After that, the obtained dataset is processed and reconstructed its shape to create a mesh model.}
\end{figure}  
\noindent The statue is fixed in the pedestal, we scan around it in different angles to get data. By another way, the scanner is fixed in a position to scan the statue on the rotated pedestal. Both of these ways can obtain 3D point clouds of the real objects, as presented in this research \cite{Sinh2019}.
\subsection{Hole filling}
This section present an important step of our proposed method. Instead of filling the holes as in previous work, in some cases, the hole is larger and it is generated on the complex location on the object's surface. Therefore, we analyze and subdivide the hole into smaller hole parts. The condition to segment at the positions of fracture margins of holes. They are determined by checking coordinates of boundary points (that is presented in detail in the previous work \cite{SinhJ2022}). In general, it is expressed as in Fig.5 below: 
\begin{figure}[H]
\center
\includegraphics[width=0.8\textwidth]{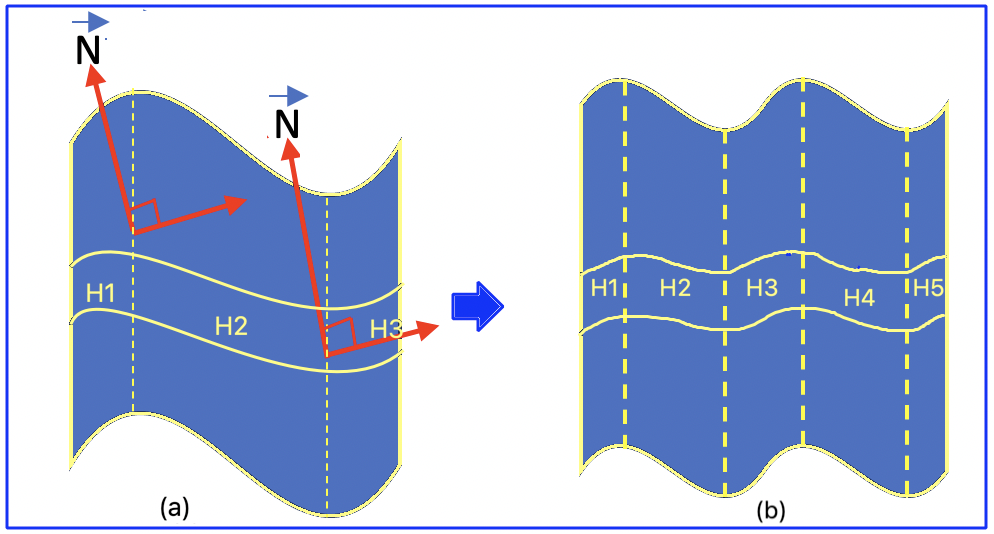}
\caption{Our proposed method for segmenting and filling on a big and complex hole.}
\end{figure}  
\noindent As in the left hand side (see Fig.5a), we determine the normal vector of the hole based on hole boundary points \cite{Sinh2016, Sinh2018} to find the position on the boundary ring for segmentation of the hole (e.g. in Fig.5a: H = H1 + H2 + H3). In a larger and more complex local curvature of the hole, we segment at the position such that each small hole is a part of the big hole with a part of local or a side of the hole (see Fig.5b). After that, we fill on each small hole (e.g. H = H1 + H2 + H3+ H4 + H5). The algorithm (Algorithm 1) is our improved step for filling a hole. 
\begin{algorithm}[h]
\caption{HoleFilling($S, H$)}
\label{alg}
\begin{algorithmic}[1]
\For {each hole $H$ on the object surface $S$}
   \State rotate the object such that the axis $Oz$ is up
   \State determine all boundary points $p_b$ of $H$
   \State compute $ds$ (the average boundary edges $e_b$)
   \State compute diameter of hole $d_H$
   \If{$d_H < 1.5$ of $ds$} //\textbf{step 1}
	  \State connect directly between the $p_b$   
   \ElsIf{$1.5 \leq d_H \leq 2.5$ of $ds$}  
      \State insert a central point $p_c$ on the hole
      \State create triangles by connecting lines($p_{b}^{i}$, $p_c$)   
   \Else
       \State determine points $p_b$ at the fracture margins
       \State segment the hole $H$ into a list of smaller holes $H_i$ //see Fig.5, Fig.6  
   \EndIf
  \For {each $H_i$}
      \State check the $d_{H_i}$
  	 \If{$d_{H_i} \leq 2.5$ of $ds$} 
  	     \State go to \textbf{step 1}
     \Else
     \For {each ring starting from the outside to inside of $H_i$} 
   		\State determine first hole boundary ring $R_f(e_b)$, process in a clock-wise order
   		\For{each $p_b$ in $R_f(e_b)$}
				\State determine its neighboring points ($p_{nei}$) in 8-connectivity 
				\If{there is any $p_{nei}$ is empty}
       				\State fill a new point by assigning $p_{new} = p_{nei}$
       				\State create a new triangle $e_b, p_{new}$
       			\EndIf
       			\State stop when the $R_f(e_b)$ is full connected
   		\EndFor	
   		\State determine the next hole boundary ring $R_n(e_b)$
   		\State $p_b \leftarrow p_{new}$
   		\State $R_f(e_b) \leftarrow R_n(e_b)$
     \EndFor
    \EndIf
   \EndFor 
\EndFor
\end{algorithmic}
\end{algorithm}
\noindent The fracture margin is determined by computing normal vector of boundary points ($\overrightarrow{p_b}$). If their normal vectors have the same direction, they are located at the same projected plane ($xoy$) of the axis $Oz$ at $p_b$ position on the boundary hole. We create a segmentation line at these boundary points to divide the hole into many sub-holes. Each part of the initial hole is then filled based on the previous work \cite{SinhJ2022}.
In order to segment the hole, we compute angle ($\cos\theta$) between the normal vectors for each pair of boundary points based on formula (4). If the $\cos\theta$ approximates 1, they are considered at the same direction. If direction of normal vector at a boundary point vary following local surface curvature of the hole and create with a normal vector of its adjacent boundary point a $\cos\theta$ greater than (in case of convex) or less than (in case of concave) a threshold value, it is determined as a segmentation point.    
\begin{equation}
	cos\theta = \frac{\vec{A}.\vec{B}}{\left \| \vec{A} \right \| \left \| \vec{B} \right \|}
\end{equation}
where \textbf{A} and \textbf{B} represents for vector $\overrightarrow{a}$ and vector $\overrightarrow{b}$ respectively.\\
After that, we compute distance between the segmentation points to connect and create a set of boundary edges for new boundary of the sub-holes. The distance between new inserted points on the segmentation line approximates the distance of $e_b$. Fig.6 illustrates in detail of our computation, segmentation and filling process (Seg1 and Seg2 are segmentation lines of the hole, at the fracture margins)
\begin{figure}[H]
\center
\includegraphics[width=\textwidth]{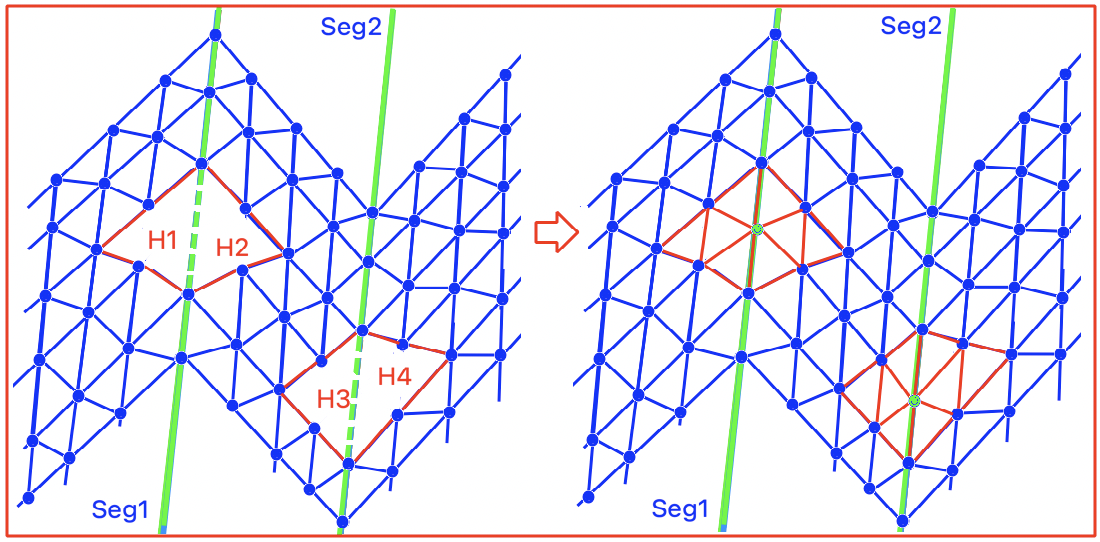}
\caption{The hole is segmented into a set of sub-holes on the left (H1, H2, H3, H4). They are filled one by one on the right.}
\end{figure}  
\noindent In order to preserve the continuity of the hole surface, we compute and adjust the z-coordinate of new inserted points adapting local curvature of the hole. As mentioned in the previous work \cite{SinhJ2022}, the z-coordinate for $p_c$ is computed as follows:
\begin{equation}
    \overrightarrow{p_c}=\frac{1}{n}\sum_{i = 1}^{n}\overrightarrow{p_{b}^{i}}	
\end{equation}
where $n$ is the number of boundary points. Similarity, z-coordinate of inserted points on the segment line ($p_{seg}$) is an average value of two segmentation points ($p_b$) on the boundary ring.
\begin{equation}
    z(p_{seg}) = \frac{z(p_{b}^{i}) + z(p_{b}^{j})}{2} 
\end{equation}
After segmenting the hole, the new points ($p_{seg}^{i}$) are inserted on the segmentation lines. They are formed to become new boundary ring for each part of the hole. The process for hole filling on each part of the hole will be performed as in previous work \cite{SinhJ2022}.

\section{Implementation and Experiment Results}
This section presents our implementation based on configuration of computer that we used to perform our proposed method. The configuration of computer includes a power CPU i7 (20 CPUs), RAM 32GB, SSD 1TB, graphical cart 4070 GeForce RTX. We use Meshlab \cite{Meshlab2023}, an open source application for processing and editing 3D triangular meshes to program and create a plugin that will perform our algorithms for reconstructing the mesh's surface. Our plugin is complied on the IDE of the Visual Studio .Net, integrated on the Meshlab. At the end, the 3D scanned object is rendered and visualized on the graphical user interface of the Meshlab application.\\
Fig.7 is an output example of a statue's surface. After triangulating its surface from point cloud data, we first detect all holes on the mesh surface; our method is then applied to reconstruct the whole objects. One by one, we process on each hole that determined on the previous step. For each hole, depends on its size and local curvature, we will segment and fill in. After filling all holes on object's surface, the reconstructed present on the right. The red color triangles on the boundary shows the boundary of hole; the pink color points shows the segmentation position of the hole.
\begin{figure}[H]
\center
\includegraphics[width=\textwidth]{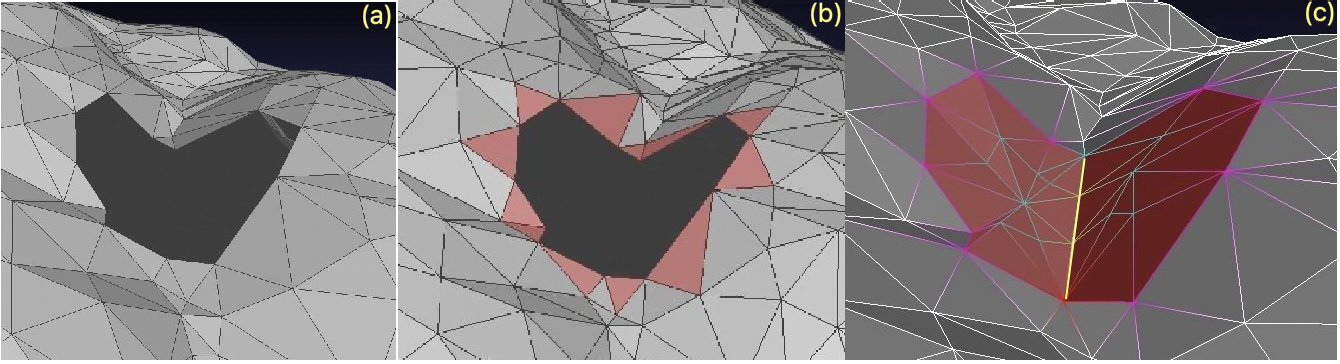}
\caption{The whole process from the left to the right: (a) a hole on the mesh surface; (b) a ring of boundary triangle is determined; (c) the hole is filled after segmenting into two parts (the yellow line is a fracture segment line of the hole).}
\end{figure}  
\noindent After reconstructing the 3D objects from real statues, we build a virtual museum that simulate for the museum of HCMC. By improving from the previous work \cite{Minh2021}, we created successfully a V-museum of HCMC \cite{Sinh2023} to illustrate. The application allow user visit the HCMC museum, interact with the statues, read their history information and also listen to the introduction from a Tour-Guide. With a VR-headset and two hand-controllers, visitors are experienced in a virtual environment that is nearly in the real museum. Fig.8 is an output example of a statue after reconstructed. We compare the accuracy between the surfaces before and after reconstructing the holes as in Table 1.
\begin{figure}[H]
\center
\includegraphics[width=\textwidth]{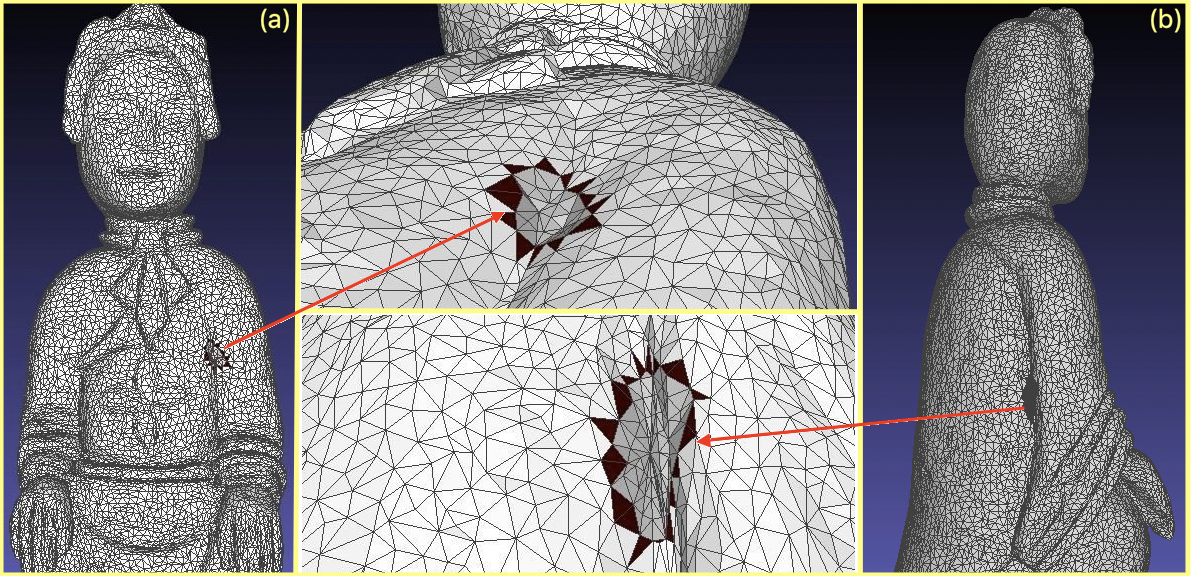}
\caption{On the left, case (a), the hole in the front of statue. On the right, case (b), the hole in the back of statue. They are filled with accuracy of the mesh surface before and after reconstructing as follows: $\Delta{max} = 0.001867$, $\Delta{avg} = 0.000001$}.
\end{figure}  

\section{Discussion and Evaluation}
In order to evaluate the quality and accuracy of the reconstructed mesh surface of the real 3D objects using our proposed method, we use the measure unit existed in the Meshlab (the Hausdorff Distance). It was performed in 2002 in the ``Metro tool" software \cite{MetroTool}, showing errors numerical values in the command prompt output. At present, the method is still very popular in the field of geometric modeling and widely used in measuring and comparing the difference between two mesh surfaces. In general, the distance $d_s$ between two surfaces ($S$ and $S'$) is defined as follows:
\begin{equation}
	d_{s}(S, S') = \underset{p \in S}{max} \: d(p, S')
\end{equation}
where $d(p, S') = \underset{p' \in S'}{min} \: d(p, p')$, and $d(p, p')$ is the Euclidean distance between two points in $R^3$ (remember that the distance is not symmetric; it means $d(S, S') \neq d(S', S)$).\\
We compute the approximation errors of the 3D object mesh surface before and after reconstructing. We measure the maximum error $\Delta_{max}$ (i.e. the two-sided Hausdorff distance), and the mean error $\Delta_{avg}$ (i.e. the area-weighted integral of the distance from point to surface) between $S$ and $S'$. The comparison of the several methods and their obtained results has been shown in Table 1 as follows:  

\begin{table}[H]
	\center
	\begin{tabular}{|c|c|c|c|c|c|c|c|}
	 	\hline
  	   The 3D&\# of &\multicolumn{2}{c|}{Close hole \cite{Meshlab2023}} & \multicolumn{2}{c|}{SSMH method \cite{SinhJ2022}} & \multicolumn{2}{c|}{\textbf{Our method}} \\
  	\cline{3-4} \cline{5-6} \cline{7-8}
  	(objects)&faces &$\Delta_{max}$ &$\Delta_{avg}$ &$\Delta_{max}$ &$\Delta_{avg}$ & $\Delta_{max}$ &$\Delta_{avg}$ \\
  	\hline
     Ba         &49924& 0.00641& 0.000002 & 0.003581 & 0.000001 & 0.00186&0.000001\\
     Chua su&     &      &    & &     &    & \\
		\hline
  	 Ba         &49719& 0.00924& 0.000006 & 0.00556& 0.000003 & 0.00271&0.000002\\
  	 Ngu hanh&     &      &    & &     &    & \\
		\hline
  	 Kim duc &49938& 0.00319&0.000001 &0.00230& 0.000001 & 0.00198&0\\
  	   TQ1    &     &      &    & &     &    & \\
  	    \hline 
  	 Kim duc&49905& 0.00544& 0.000002 &0.00522&0.000002  & 0.00275&0.000001\\	
  	    TQ2 &     &      &    & &     &    & \\
  	   \hline
  	 Ong   &49929& 0.00412&0.000002 & 0.00226&0.000001 & 0.00200& 0.000001\\	
  	  Dia    &     &      &    & &     &    & \\
  	    \hline
  	 Phuc duc&49965& 0.00305& 0.000001 & 0.00272& 0 &0.00132& 0\\	
  	    CT     &     &      &    & &     &    & \\
  	   \hline
  	  Quan &49767& 0.00532& 0.000004 & 0.00368& 0.000003& 0.00238&0.000002\\	
  	  Cong &  & &  &    &  &   &\\
       \hline
  	  Thuy duc&49855&0.00303&0.000001 & 0.00253&0.000001 & 0.00151&0.000001\\	
  	    TQ &  & &  &    &  &   &\\
  	   \hline  
	\end{tabular}  
  \caption{Comparing approximation errors ($\Delta_{max}$ and $\Delta_{avg}$) of the reconstructed objects among the methods after filling all holes on the surface of each statue. Close hole \cite{Meshlab2023} is a method integrated on Meshlab. The SSMH method is used in the previous work \cite{SinhJ2022}. Our method is proposed in this research.}
\end{table}
\noindent Fig.9 indicate a case of more than one holes on the object's surface. After determining and filling all the holes based on our proposed method, the accuracy of the reconstructed surface is approximating its original model.
\begin{figure}[H]
\center
\includegraphics[width=\textwidth]{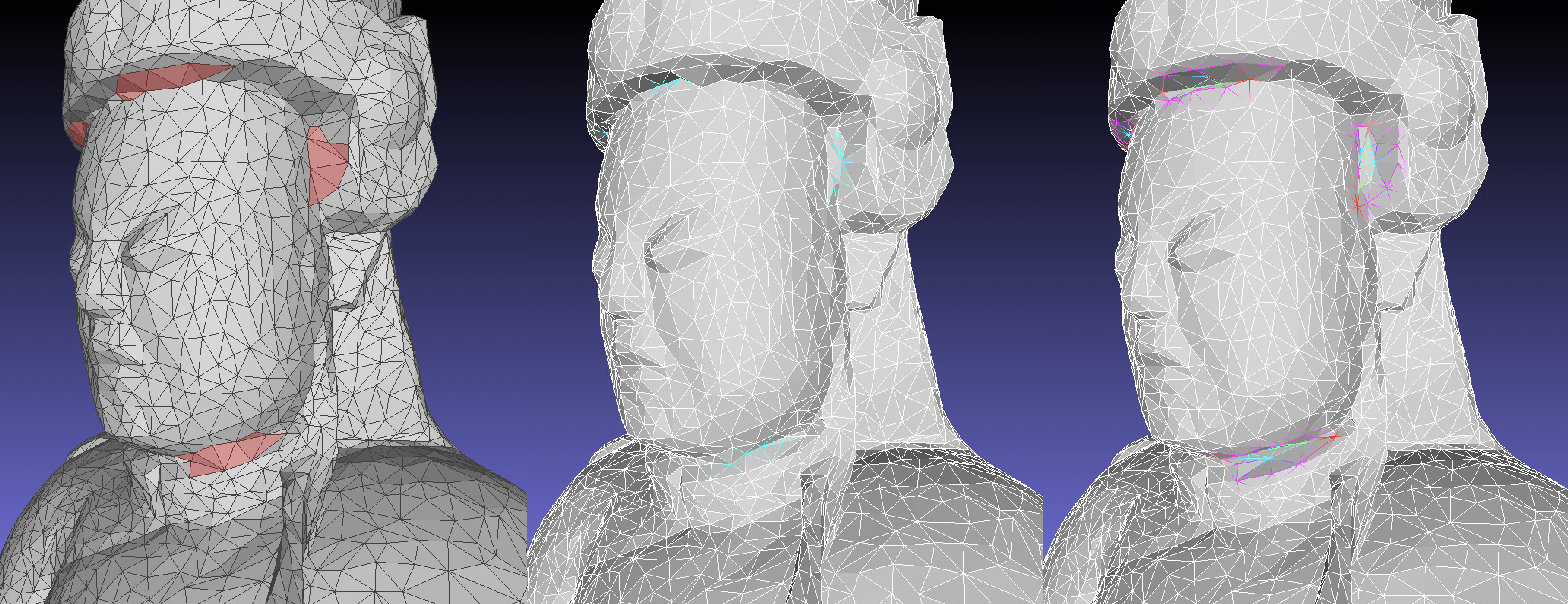}
\caption{The holes on the statue (with 49754 faces) are filled by using three methods, from the left to right: Close hole method \cite{Meshlab2023} ($\Delta{max} = 0.005713$, $\Delta{avg} = 0.000002$). SSMH method in the previous work \cite{SinhJ2022} ($\Delta{max} = 0.004014$, $\Delta{avg} = 0.000002$). Our proposed method ($\Delta{max} = 0.002247$, $\Delta{avg} = 0.000001$).}
\end{figure}  
\noindent Fig.10 visualizes all statues that we have processed, reconstructed and built in a virtual of HCMC museum. This application can be accessed our demo version at https://www.youtube.com/watch?v=LJQ2LUYDrOU \cite{Sinh2023}.
\begin{figure}[H]
\center
\includegraphics[width=\textwidth]{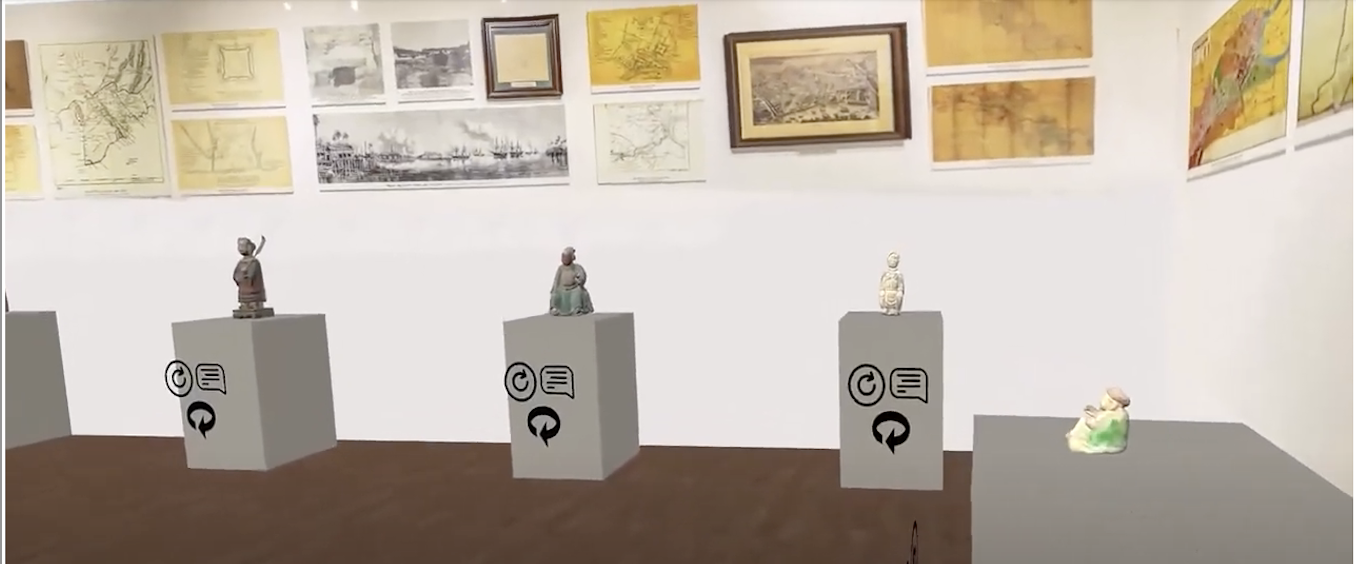}
\caption{Our application of HCMC V-museum}.
\end{figure}  
\noindent To show more clearly the improved point of our proposed method, we add one more picture to compare directly the obtained result of images after using three hole filling methods (see Fig.11). The Close hole method process very fast. It connect directly between the boundary points without inserting any points. Therefore, the hole is not followed local surface curvature of the hole and the accuracy is low. The SSMH method is more exact comparing to the Close hole method. However, comparing to our proposed method, the accuracy is a little bit lower and the hole patch is not adapt exactly local curvature as our proposed method.
\begin{figure}[H]
\center
\includegraphics[width=\textwidth]{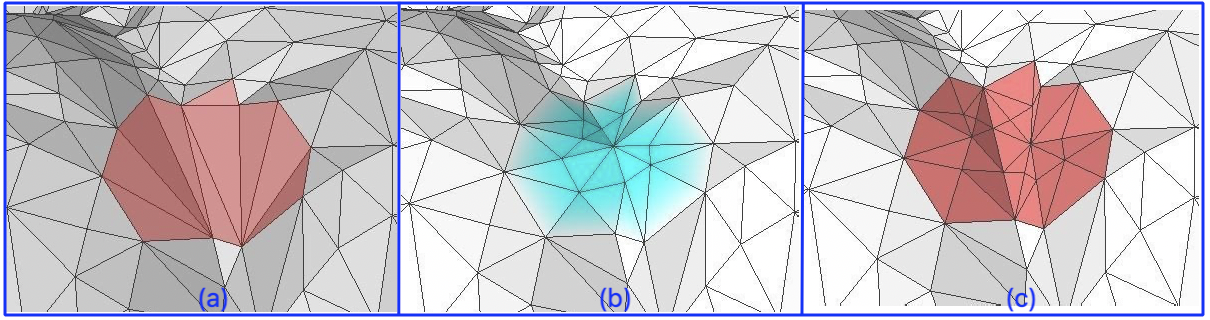}
\caption{Comparing the reconstructed model by filling the hole based on three methods: (a) Close hole method, \cite{Meshlab2023} ($\Delta{max} = 0.004678$, $\Delta{avg} = 0.000018$). SSMH method in the previous work \cite{SinhJ2022} ($\Delta{max} = 0.004189$, $\Delta{avg} = 0.000022$). Our proposed method ($\Delta{max} = 0.003678$, $\Delta{avg} = 0.000018$).}.
\end{figure}

\section{Conclusions}
In this research, we presented our method for reconstructing the 3D real objects of the heritage statues. The input data are 3D point clouds of the statues that is scanned in the museum of HCMC. The holes come from status of the real objects and are generated on the surface of the real objects during scanning step. After processing the obtained dataset, removing noisy data, we triangulated its surface. The holes on the surface are then filled using our proposed method. The novelty point of our method is processing directly on a larger and complex hole on the object's surface. We analyze the structure of the hole, determine the position on the hole boundary to segment it depending on fracture margins of holes. Each part of the holes is then filled by applying the method in the previous work \cite{SinhJ2022}. The proposed method obtained higher accuracy, the reconstructed surface is close to its shape and approximate the initial curvature. Comparing to previous work, we have improved the filling step. After filling the hole, the real object is restored nearly completely for building successfully a V-museum application. 
\section{Acknowledgments} 
This research is funded by Vietnam National University HoChiMinh City (VNU-HCM) under grant number DS2023-28-01. We would like to thank for the fund.


\begin{thebibliography}{9}
\bibitem{Jonh2019}Jonh Vince. Mathematics for Computer Graphics. Fifth Edition, ISBN: 978-1-4471-7334-2, Springer Nature 2019.
\bibitem{Steve2021}Steve Marschner, Peter Shirley. Fundamentals of Computer Graphics, Fifth Edition. Taylor \& Francis Group, LLC, 2022.
\bibitem{Minh2021}Minh Khai Tran, Sinh Van Nguyen, Nghia Tuan To, Marcin Maleszka. Processing and Visualizing the 3D Models in Digital Heritage. 13th International Conference on Computational Collective Intelligence (ICCCI 2021). Rank B. Lecture Notes in Computer Science, vol 12876. Pages. 613-625. https://doi.org/10.1007/978-3-030-88081-1\_46, Springer 2021.
\bibitem{revos2023}Metrology, R-EVO S-SCAN Articulated measuring arm with laser scanner. https://www.rpsmetrology.com/en/product/r-evo-s-scan/. Last accessed Jan 2023.
\bibitem{SinhJ2022}Nguyen Van Sinh, Le Thanh Son, Tran Khai Minh, Tran Man Ha. Reconstruction of 3D digital heritage objects for VR and AR applications. Journal of Information and Telecommunication. ISSN: 2475-1839. Vol. 6, No. 3, Pages. 254-269. Taylor \& Francis, doi.org/10.1080/24751839.2021.2008133. 2021.
\bibitem{hcmmuseum}Museum of Ho Chi Minh City of Vietnam. https://hcmc-museum.edu.vn/en/trang-chu-english/ Accessed May 2023.
\bibitem{NVS2013}NGUYEN Van-Sinh (2013). 3D Modeling of elevation surfaces from voxel structured point clouds extracted from seismic cubes. PhD Thesis, Aix-Marseille University.
\bibitem{Marco2012}Marco Abate, Francesca Tovena. Curves and Surfaces. ISSN: 2038-5722, Springer 2012.
\bibitem{Penev2013}Alexander Penev (2013). Computer Graphics and Geometric Modelling - A Hybrid Approach. International Journal of Pure and Applied Mathematics. ISSN 1311-8080, Volume 85, No. 4, Pages. 781-811.
\bibitem{David2001}David F.Rogers. An Introduction To NURBS With Historical Perspective. ISBN-10:1-55860-669-6, Academic Press, 2001.
\bibitem{Sinh2018}Nguyen V.S., Tran K.M. and Tran M.H (2018). Filling Holes on The Surface of 3D Point Clouds Based on Reverse Computation of Bezier Curves. Information Systems Design and Intelligent Applications. Advances in Intelligent Systems and Computing. Volume 672, ISSN: 2194-5357, Pages 334-345, Springer.
\bibitem{Licio2013}Licio H.B (2013). Efficient computation of Bezier curves from their Bernstein-Fourier representation. Journal of Applied Mathematics and Computation, Vol.220, pp.235-238.
\bibitem{SHM2018}N.V. Sinh, T.M. Ha and T.K. Minh (2018). An Improved Method for Restoring The Shape of 3D Point Cloud Surfaces. International Journal of Synthetic Emotions (IJSE), ISSN: 1947-9093, EISSN: 1947-9107, Vol 9, Issue 2, Article 3, Pages: 37-53.
\bibitem{Ergun2021}Ergun, Bahadir \& Şahin, Cumhur. (2021). Laser Point Cloud Segmentation in MATLAB. 10.5772/intechopen.95249. 
\bibitem{Luca2015}Luca Di Angelo and Paolo Di Stefano. Geometric segmentation of 3D scanned surfaces. Computer-Aided Design 62 (2015), pp 44–56. 2015.
\bibitem{Gen2008}Gen Li, Xiu-Zi Ye, San-Yuan Zhang. An algorithm for filling complex holes in reverse engineering. Engineering with Computers (2008) 24:119–125. DOI 10.1007/s00366-007-0075-9
\bibitem{Luca2023}Calatroni, L., Huska, M., Morigi, S. et al. A Unified Surface Geometric Framework for Feature-Aware Denoising, Hole Filling and Context-Aware Completion. Journal of Mathematical Imaging and Vision (2023) 65:82–98, https://doi.org/10.1007/s10851-022-01107-w
\bibitem{Sinh2013}Nguyen Van-Sinh, Bac Alexandra and Daniel Marc (2013). Simplification of 3D Point Clouds sampled from Elevation Surfaces, 21st International Conference on Computer Graphics, Visualization and Computer Vision WSCG. pp.60-69, ISBN:978-80-86943-75-6.
\bibitem{Chunhong2017}Chunhong X. and Hui Z (2017). A fast and automatic hole-filling method based on feature line recovery, Journal of Computer-Aided Design and Applications. Vol. 14(6), pp. 751-759.
\bibitem{Sinh2014}V.S. NGUYEN, A. BAC, M. DANIEL (2014). Triangulation of an elevation surface structured by a sparse 3D grid, The Fifth IEEE International Conference on Communications and Electronics IEEE ICCE 2014, IEEE ISBN: 978-1-4799-5049-2, Pages 464-469.
\bibitem{Sinh2019}NGUYEN Van Sinh, TRAN Manh Ha, LE Quang Minh Anh (2019). A Research for Reconstructing 3D Object by Using an RGB-D Camera, Frontiers in Intelligent Computing: Theory and Applications. Advances in Intelligent Systems and Computing (AISC), vol 1014, Chapter 2, Pages. 13-24, ISBN: 978-981-13-9919-0. Springer.
\bibitem{Sinh2016}Nguyen V.S., Tran M.H. and Nguyen T.T (2016). Filling Holes on The Surface of 3D Point Clouds Based on Tangent Plane of Hole Boundary Points. The Seventh International Symposium on Information and Communication Technology (SoICT). ACM ISBN: 978-1-4503-4815-7, Pages 331-338.
\bibitem{SinhJ2021}Sinh Van Nguyen, Ha Manh Tran, Mercin Maleszka. Geometric Modeling: Background for Processing the 3D Objects. Applied Intelligence, ISSN: 1573 - 7497. Vol.51, No.8,Pages. 6182–6201. (SCI-E, Q2, IF: 5.08), Feb, 2021.
\bibitem{Gou2022} Gou, G.; Sui, H.; Li, D.; Peng, Z.; Guo, B.; Yang, W.; Huang, D. LIMOFilling: Local Information Guide Hole-Filling and Sharp Feature Recovery for Manifold Meshes. RemoteSens.2022,14,289. https:// doi.org/10.3390/rs14020289
\bibitem{Jules2018}Jules Morel, Alexandra Bac, Cedric Vega. Surface reconstruction of incomplete datasets: A novel Poisson Surface approach based on CSRBF. Computer \& Graphics, 74(2018), pp.44-55. https://doi.org/10.1016/j.cag.2018.05.004, 2018. 
\bibitem{Guo2016}Guo, Xiaoyuan and Xiao, Jun and Wang, Ying. A survey on algorithms of hole filling in 3D surface reconstruction.  Journal of The Visual Computer.  Vol 34(1), pages 93-103, Doi 10.1007/s00371-016-1316-y. 2016.
\bibitem{Perez2016}E., Salamanca S., Merchán P. and Adan A (2016). A comparison of hole-filling methods in 3D. International Journal of Applied Mathematics and Computer Science. ISSN (Online) 2083-8492, Volume 26(4), Pages 885-903.
\bibitem{Custodio2023}A. L. Custódio  and M. A. Fortes  and A. M. Sajo-Castelli. Filling holes under non-linear constraints. Journal of Computational and Applied Mathematics (2023) 42:72 https://doi.org/10.1007/s40314-023-02210-3  
\bibitem{Cui2021}Cui,L.;Zhang,G.;Wang,J. Hole Repairing Algorithm for 3D Point Cloud Model of Symmetrical Objects Grasped by the Manipulator. Sensors2021,21,7558. https:// doi.org/10.3390/s21227558
\bibitem{Bin2021}Bin Liu, Mingzhe Wang, Xiaolei Niu, Shengfa Wang, Song Zhang, Jianxin Zhang. A Fragment Fracture Surface Segmentation Method Based on Learning of Local Geometric Features on Margins Used for Automatic Utensil Reassembly. Computer-Aided Design 132 (2021) 102963, pp 1-15, 2021.
\bibitem{Du2023}Du, W., Xia, Z., Han, L. et al. Correction to: 3D solid model generation method based on a generative adversarial network. Appl Intell 53, 18107 (2023). https://doi.org/10.1007/s10489-023-04459-x
\bibitem{Weizhi2019}Weizhi Nie, Weijie Wang, Anan Liu, Yuting Su and Jie Nie (2019) HGAN: Holistic Generative Adversarial Networks for Two-dimensional Image-based Three-dimensional Object Retrieval. ACM Trans. Multimedia Comput. Commun. Appl., Vol. 15, No. 4, Article 101.
\bibitem{Alok2023}Jhaldiyal, A., Chaudhary, N. Semantic segmentation of 3D LiDAR data using deep learning: a review of projection-based methods. Appl Intell 53, 6844–6855 (2023). https://doi.org/10.1007/s10489-022-03930-5.
\bibitem{Chen2022}Chen, C., Wang, Y., Chen, H. et al. GeoSegNet: point cloud semantic segmentation via geometric encoder-decoder modeling. Vis Comput (2023). https://doi.org/10.1007/s00371-023-02853-7. 2023
\bibitem{Meshlab2023}Meshlab (access 2023). Institute of the National Research Council of Italy CNR. http://www.meshlab.net.
\bibitem{Sinh2023} Sinh Van Nguyen, Duy Bao Dinh, Son Thanh Le, Sach Thanh Le, Lam Son Quoc Pham, Marcin Maleszka and Lam Duc Vu Nguyen. A Solution for Building a V-Museum Based on Virtual Reality Application. International scientific conference for research in the field of Computational Collective Intelligence (ICCCI 2023, Budapest Hungary). Accepted to present.
\bibitem{MetroTool}Cignoni, Paolo \& Rocchini, Claudio \& Scopigno, Roberto. METRO: Measuring error on simplified surfaces. Computer Graphics Forum. 17. 167 - 174. 10.1111/1467-8659.00236. 1998.
\end{thebibliography}
\end{document}